\documentclass[10pt,journal,compsoc]{IEEEtran}
\usepackage{multirow} %for tables again
\usepackage{bigstrut} %tables
\usepackage{algorithm}
\usepackage{enumitem}
\usepackage{newfloat}
\usepackage{hyperref}
\usepackage{listings}
\usepackage{algorithmic}
\usepackage{amsmath} %for writing maths
\usepackage{adjustbox}%supplement to graphics package
\usepackage{multirow}
\usepackage{graphicx}
\usepackage{subcaption}
\usepackage{lscape} %allows
\ifCLASSOPTIONcompsoc
  \usepackage[nocompress]{cite}
\else
  \usepackage{cite}
\fi
\ifCLASSINFOpdf
\else
\fi

\usepackage{graphicx}
\begin{document}
\title{TagRec++: Hierarchical Label Aware Attention Network for Question Categorization}


\author{Venktesh V,
        Mukesh Mohania,
        and~Vikram~Goyal% <-this % stops a space
\IEEEcompsocitemizethanks{\IEEEcompsocthanksitem Venktesh V, Mukesh Mohania and Vikram Goyal are with Dept. of CSE, IIIT-Delhi, India\protect\\
% E-mail: see http://www.michaelshell.org/contact.html
.}% <-this % stops an unwanted space
% \thanks{Manuscript received April 19, 2005; revised August 26, 2015.}
}


\markboth{IEEE Transactions on Knowledge and Data Engineering}%
{Shell \MakeLowercase{\textit{et al.}}: Bare Demo of IEEEtran.cls for Computer Society Journals}



\IEEEtitleabstractindextext{%
\begin{abstract}
Online learning systems have multiple data repositories in the form of transcripts, books and questions. To enable ease of access, such systems organize the content according to a well defined taxonomy of hierarchical nature (subject - chapter -topic). The task of categorizing inputs to the hierarchical labels is usually cast as a flat multi-class classification problem. Such approaches ignore the semantic relatedness between the terms in the input and the tokens in the hierarchical labels. Alternate approaches also suffer from class imbalance when they only consider leaf level nodes as labels. To tackle the issues, we formulate the task as a dense retrieval problem to retrieve the appropriate hierarchical labels for each content. In this paper, we deal with categorizing questions. We model the hierarchical labels as a composition of their tokens and use an efficient cross-attention mechanism to fuse the information with the term representations of the content. We also propose an adaptive in-batch hard negative sampling approach which samples better negatives as the training progresses. We demonstrate that the proposed approach \textit{TagRec++} outperforms existing state-of-the-art approaches on question datasets as measured by Recall@k. In addition, we demonstrate zero-shot capabilities of \textit{TagRec++} and ability to adapt to label changes. 
\end{abstract}

}

\maketitle

\IEEEdisplaynontitleabstractindextext

% \fi
\IEEEpeerreviewmaketitle

\IEEEraisesectionheading{\section{Introduction}\label{sec:introduction}}

Online educational systems organize content like questions according to a hierarchical learning taxonomy of form subject-chapter-topic. For instance a content about ``pH level" is tagged with the learning taxonomy \textit{"science - chemistry - acids"}. In the above example \textit{science} is the root node and \textit{acids} is the leaf node. Organization of content in such standard format aids in better accessibility as users can easily navigate through large repositories of learning content by searching using different \textit{facets} like the subject, chapter or topic names. The automated taxonomy tagger would aid in on-boarding content at scale from other sources by tagging them with a standardized learning taxonomy. However, manual labeling of content with the hierarchical taxonomy is cumbersome. Automated tagging of content with learning taxonomy would enable indexing of content at scale and conserve time. 

The class of problems involving categorization of content to labels of hierarchical form are usually cast as flat multi-class classification tasks \cite{xumulti,kozareva2015everyone}. The flat classification approaches ignore the hierarchical structure in the label space and encode the labels as numbers. Several approaches \cite{flat1, flat2, flat3} consider only the leaf nodes as labels to reduce the label space. In the former method the hierarchy is ignored and in the latter the problem of class imbalance occurs as most of the content is attached to a few leaf nodes. Another challenge is the \textit{open-set identification} problem where new labels may emerge in the label space owing to addition of new topics or removal of old topics. The new hierarchical labels would still be semantically related to the old labels and hence the model must be able to adapt to changes in the label space without re-training. Further, traditional multi-label multi-class classification approaches require changes in the model architecture and re-training to adapt to changes in the label space.

\begin{figure*}
    \centering
    \includegraphics[width =1\linewidth]{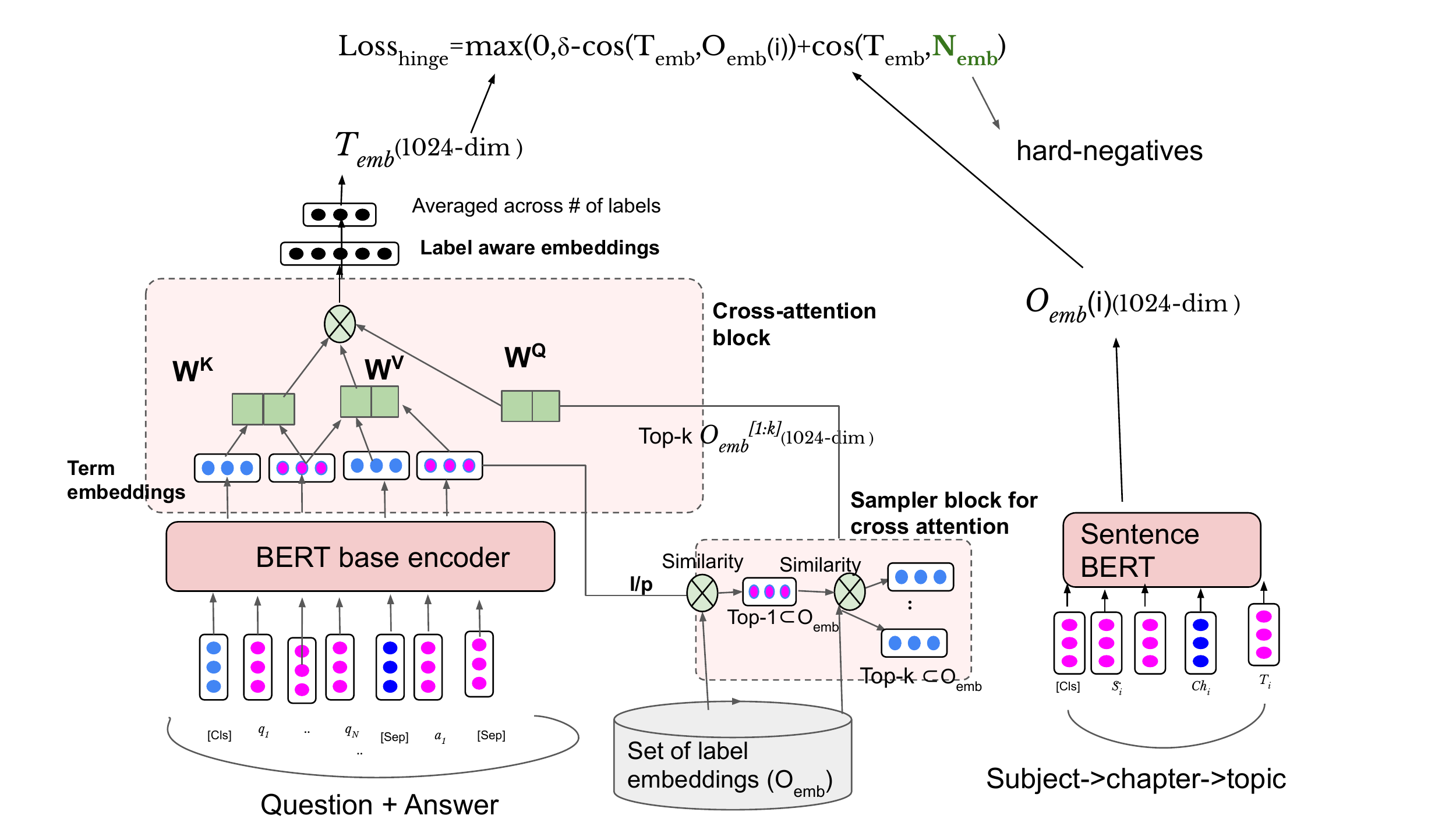}
    \centering \caption{Proposed approach: TagRec++}
    \label{tagrec++}
\end{figure*}

The proposed approach, \textit{TagRec++}, can accommodate changes in label space without re-training. The problem is viewed as a dense retrieval task \cite{twinbert} where the goal is to retrieve the most relevant labels for a given question as shown in Figure \ref{tagrec++}. Since the tokens in the label are abstractions of their word descriptions, they are related to the terms in the content. Hence, we adopt a contrastive learning approach to project the content and labels to a continuous vector space and ensure the label representations are aligned with their corresponding content representations. We extend upon the work of TagRec \cite{Tagrec} which proposed a simple two-tower architecture for bringing together the appropriate vector sub-spaces of the input and hierarchical labels closer. The \textit{TagRec++} fuses the content representations with the related hierarchical labels using an interactive attention mechanism. It helps to better capture the relationship between the terms in the input learning content and the tokens in the hierarchical taxonomy. Specifically it uses a late interaction approach where the label embeddings can be pre-computed and indexed unlike cross-encoder based approaches. The \textit{TagRec++} uses a contrastive learning approach where triplets (anchor, positive, negative) are mined to pull apart the negatives from the anchor and bring the anchor and the positive sample closer. The sampling of negatives helps in learning better representations that have higher capacity to distinguish between positive and negative labels thereby increasing recall during retrieval. Hence, \textit{TagRec++} uses an adaptive in-batch hard-negative sampling approach where hierarchical labels closer to the content representation are chosen as hard-negatives. These labels do not correspond to the ground truth for the content yet their representations are closer to the content representations. They are sampled dynamically in the training loop. Hence, as the model parameters are updated during training, it improves at the task of sampling harder negatives further helping in disentangling the vector space of positive and negative labels. This aids in aligning the input representations with the appropriate label representations.

The \textit{TagRec++} is evaluated on datasets of question banks in science and related domains. The content may have both the question and its answer. If the answer is given we combine the question with the answer to capture more semantic context. The combined "question-answer" or the "question" in isolation is our learning content and we will use the terms "content" and "question" interchangeably in the rest of the discussion. To further demonstrate that the \textit{TagRec++} approach is general for the taxonomy tagging task on diverse content, it is evaluated on a dataset of MCQs (Multiple Choice Questions) with no semantic context in answers. Further we also evaluate it for the task of tagging related short content like learning objectives in a zero-shot setting.

The \textit{TagRec++} can be used in several applications. Tagging content to a standardized taxonomy can be used for redirection of questions to relevant subject matter experts in question answer forums of such systems. The questions once tagged to a standardized taxonomy can be redirected to known experts according to the topic tags. It can be used in forums like \textit{Stackoverflow}. For instance a question about \textit{Haskell} can be tagged with \textit{Computer Science$\xrightarrow{}$functional programming$\xrightarrow{}$Haskell}. Another application of categorizing content to a standardized learning taxonomy in such systems is the automated linking of diverse learning content. For instance, related questions and reading material can be linked by tagging them to a learning taxonomy. This helps reduce the search space when recommending related reading material to a student for solving a question.

The core contributions of our work are:
\begin{itemize}
    \item We propose the task of automated content categorization involving hierarchical labels as a dense retrieval problem and adopt a contrastive learning approach.
    
    \item We propose an interactive attention approach between the content representations and the hierarchical labels to fuse the information in the hierarchical labels with the input representations for better learning and convergence. We render this process efficient by grouping related labels to reduce the label space and provide bounds for the approximation leveraging Lipschitz continuity principle.
    
    \item We propose an adaptive in-batch hard negative sampling approach in the training loop of the contrastive learning approach for high recall retrieval.
    
    \item We experimentally study the proposed approach on datasets with different characteristics and demonstrate its superior performance compared to existing methods as measured by Recall@k. We also provide an analysis of the model's ability to adapt to changes in label space. The study also demonstrates the ability of our approach to tag related content like learning objectives in a zero-shot setting.
    
\end{itemize}

The rest of the paper is organized as follows:
\begin{itemize}
    \item We present a literature review of text classification methods involving labels of hierarchical structure in Section \ref{related}.
    \item We present in detail the proposed approach in Section \ref{methods}. We explain the various components of the proposed approach.
    \item In Section \ref{experiments}, we describe the data preparation method, experimental setup, baselines and the ablation studies performed.
    
    \item We present an analysis of results in Section \ref{results}.
    \item We conclude with scope for future work in Section \ref{conclusion}.
\end{itemize}
\textbf{Reproducibility}: To ensure reproducible work, we open-source our code and datasets at \url{https://github.com/ADS-AI/TagRec\_Plus\_Plus\_TKDE}.


\section{Related Work}
\label{related}
In this section, we provide an overview of approaches that tackle problems involving labels of hierarchical nature. We also give an overview of state-of-the-art vector representation methods.

\subsection{Text categorization to Hierarchical Labels}
The online systems use a standardized taxonomy to organize their content \cite{xumulti,kozareva2015everyone}. The taxonomy is of hierarchical nature and usually, the approaches used to categorize content in such taxonomy can be categorized into multi-class classification or hierarchical multi-step approaches \cite{article,yu2012product}. In multi-class single-step methods, the leaf nodes are considered labels while ignoring the hierarchy. This leads to class imbalance issue where a large number of samples cover only a few leaf nodes considered as labels. In the hierarchical multi-step approach, the root category is predicted using a classifier and the process is repeated to predict the nodes at the subsequent level. However, the main issues of the approach are that the number of classifiers increases with depth, and the error from one level propagates to the next level. This also increases the computation needed at the inference time. Along similar lines, Banerjee et al. \cite{banerjee-etal-2019-hierarchical} proposed to build a classifier for each level. However, unlike the previous works, the parameters of the classifier for parent levels are transferred to the classifiers at child levels. Another approach \cite{chained_neural_networks} proposed to use a chain of neural networks to categorize content to hierarchical labels. A classifier is designated for each level in the hierarchy. However, the major limitation here is that the number of networks in the chain increases with depth, and it also requires that the paths in the label hierarchy should be of the same length, limiting the applications to cases of minor changes in the label space.  

To circumvent these issues, each hierarchical taxonomy could be considered as a label disregarding the hierarchy, and a regular single-step classifier could be trained to classify the content to one of the labels.
Several single-step classifiers have been proposed for classification tasks involving hierarchical labels. In \cite{yu2012product}, the word level features like n-grams were used with SVM as a classifier to predict level 1 categories, whereas in \cite{kozareva2015everyone} the authors have leveraged n-gram features and distributed representations from Word2Vec followed by a linear classifier for multi-class classification. Several deep learning methods like CNN \cite{10.1145/3302425.3302483} and LSTM \cite{article} have been proposed for the task of question classification. Since the pre-trained language models, like BERT \cite{BERT}, improve the performance, the authors in \cite{xumulti} propose a model BERT-QC, which fine-tunes BERT to categorize questions from the science domain to labels of hierarchical format. The problems involving hierarchical labels have also been formulated as a translation problem in \cite{MachinT} where the product titles are provided as input and use a seq2seq architecture to translate them to product categories having hierarchical structure. The hierarchical neural attention model \cite{sinha2018hierarchical}  leverages attention to obtain useful input sentence representation and uses an encoder-decoder architecture to predict each node in the hierarchical learning taxonomy. However, this approach may not scale as the depth of the hierarchy increases.

Several works have also tried to capture the hierarchical structure of the labels for the purpose of text categorization to such labels. For instance, Zhou et al. \cite{zhou-etal-2020-hierarchy} proposed to design an encoder that incorporates prior knowledge of label hierarchy to compute label representations. However, they flatten the hierarchy and treat every label as a leaf node which would require re-training when there are changes in the label space. Lu et al. \cite{lu-etal-2020-multi} introduced different types of label graphs (co-occurrence based and semantic similarity based) to improve text categorization. However, they also cast the task of categorizing text to hierarchical labels as a multiple binary classification task \cite{Frnkranz2008MultilabelCV}. These approaches do not consider the relationship between the terms in the text inputs and the hierarchical labels.

\subsection{Sentence representation methods}
The NLP tasks like classification and retrieval have been advanced by distributed representations that capture the semantic relationships \cite{10.5555/2999792.2999959}. Methods like GloVe \cite{pennington-etal-2014-glove} compute static word embeddings which do not consider the context of occurrence of the word. An unsupervised method named Sent2Vec \cite{pagliardini2017unsupervised} was proposed to create useful sentence representations.

The Bidirectional Encoder Representation from Transformers (BERT) \cite{BERT} does not compute any independent sentence representations.
To tackle this, Sentence-BERT \cite{reimers-gurevych-2019-sentence} model was proposed to generate useful sentence embeddings by fine-tuning BERT. Another transformer-based sentence encoding model is the Universal Sentence Encoder (USE) \cite{cer-etal-2018-universal}  that has been specifically trained on Semantic Textual Similarity (STS) tasks and generates useful sentence representations.

\setlength{\parindent}{4ex}In this paper, sentence representation methods are used to represent the labels by treating each of them as a sequence. For example, the label \textbf{Science - Physics - alternating current} is treated as a sequence. We employ sentence representation methods to model the compositionality of terms present in the hierarchical learning taxonomy. Several works \cite{compositionality1} \cite{compositionality2} have demonstrated that sentence representation models are able to capture the nature of how terms compose together to form meaning in a sequence. We posit that the same principle can be used to capture the complete semantic meaning of hierarchical taxonomy in the vector space. In our experiments, we observe that sentence transformer based representation methods perform better than averaging word embeddings.
\subsection*{Difference between TagRec \cite{Tagrec} and TagRec++}
In the work TagRec\cite{Tagrec}, the authors take a dense retrieval approach with the aim to retrieve the relevant label (i.e., the hierarchical learning taxonomy) by aligning the input embeddings and the label embeddings. The authors propose a bi-encoder based architecture \cite{mazare-etal-2018-training} \cite{biencoder2} to accomplish the task of aligning the inputs with the correct hierarchical labels. Though the input and label representations are matched using a hinge ranking loss, the proposed approach doesn't explicitly combine the information from the input and label spaces. We posit that explicitly capturing the relationship between the terms in the input and the hierarchical labels can help in high recall retrieval. Hence, we propose an interactive attention mechanism where the related labels closer to the input are grouped and attention is computed with respect to all terms in the input to produce hierarchical taxonomy aware representations. In the work TagRec \cite{Tagrec}, the authors use a random in-batch negative sampling \cite{randomneg} for the ranking loss in the contrastive learning setup. However, in dense retrieval, it is common knowledge \cite{hardneg1}, \cite{hardneg2} that hard negatives aid in high recall retrieval. However, the existing approaches usually leverage a warm start Dense Retrieval model to build a cache of hard negatives and constantly refresh them, adding to computational cost. They also do not refresh the hard negatives based on the parameters of the model being trained. On the other hand, this work proposes to use a dynamic in-batch hard negative sampling approach that provides better negatives as the model parameters are updated in the training loop. We also demonstrate that it outperforms the in-batch random negatives commonly deployed in dense retrieval tasks.

In this work, the multi-class classifier approach is not explored owing to the limitations explained earlier in this section. The proposed method \textit{TagRec++} can also adapt to changes in the label space and demonstrates impressive zero-shot performance on related short text content.

\section{Methodology}
\label{methods}
In this section, we describe our method for retrieving relevant hierarchical labels for learning content documents, with each document corresponding to a question-answer pair. The architecture for the proposed approach is as shown in Figure 1. The proposed approach \textit{TagRec++} aligns the input representations with the corresponding hierarchical labels representations. It is composed of an interactive cross-attention mechanism that fuses information from the hierarchical labels and the input to render taxonomy aware representations. It also uses an in-batch hard negative sampling approach to pull apart negative labels and bring the input representations closer to the positive label representations.

The input to the method is a corpus of documents, $C=\{D_1,D_2...D_n\}$. The documents are tagged with hierarchical labels $O=\{(S_1,Ch_1,T_1),(S_2,Ch_2,T_2)... | S_i > Ch_i > T_i \}$ where $S_i$ (root node),
$Ch_i$ and $T_i$ (leaf node) denote subject, chapter, and topic, respectively. The goal here is to learn an input representation that is close in the continuous vector space to the correct label representation. We consider the label $(S_i,Ch_i,T_i)$ as a sequence, ($S_i+Ch_i+T_i$) and obtain a sentence representation for it using pre-trained models. Since the sentence representation encoder is frozen during training, we can precompute and index the embeddings for the hierarchical labels. This also ensures faster inference. We leverage BERT \cite{BERT} for obtaining contextualized embeddings followed by a linear layer. The linear layer maps the 768-D representation from BERT to the 1024-D or 512-D vector representation. The novelty of the method lies in the implementation of the attention layer and in-batch hard negative sampling, which are discussed in detail later in the Section.  

The steps of the proposed approach can be observed from Algorithm 1. During the \textit{training phase}, the input text is cast to a continuous vector space using a BERT \cite{BERT} base model. The labels are projected to a continuous vector space using a Sentence-BERT model to capture the composition of terms in the label. To capture the interaction of terms in the content with hierarchical labels, we fuse the information through an attention mechanism between the embeddings of related hierarchical labels and the term embeddings of the content. Then we use a hinge rank-based loss to align the input vector representations with the correct label representations. We also propose an in-batch hard-negative sampling approach to pull apart irrelevant labels for a given input content.
 \begin{algorithm}
 \caption{Tag Recommender}
 \label{algoBO}
 \begin{algorithmic}[1]
 \renewcommand{\algorithmicrequire}{\textbf{Input:}}
 \renewcommand{\algorithmicensure}{\textbf{Output:}}
 \REQUIRE Training set $T \gets $ docs $\{D_1,..D_n\}$, labels $O$ of form (Subject-Chapter-Topic)\\
 \ENSURE  Set of tags for test set , $RO$
  \\ \textbf{Training} (batch mode)
  \STATE Get input text embeddings , $T_{emb} \gets BERT(D)$
  \STATE Obtain label embeddings, $O_{emb} \gets SENT\_BERT(O)$ 

  \STATE $Index(labels) \gets O_{emb}$
    \STATE Get label embedding closest to input, \\ $O_{emb}^i \gets top_1(cos(T_{emb},Index(labels)))$
  \STATE Get top labels close to selected label, \\ $L_r \gets ||O_{emb}^i - index(labels)||_2$ 
  
  \STATE Project labels to queries, $Q \gets W^Q * L_r$
  
  \STATE Project input embeddings to key and values, \\ $K \gets W^K * T_{emb}$ and $V \gets W^V * T_{emb}$
  
  \STATE Compute modified input embeddings, \\ $T_{new} \gets \dfrac{Softmax(Q * K^T)}{\sqrt{d_k}} *V $
  
  \STATE $hard\_neg \gets top\_k(cos(T_{new},Index(labels)))$ excluding $label$ (adaptive hard-negatives)
    \STATE          $L(text,l)\gets \sum_{j \neq l}max(0,\delta-cos(T_{new},v(l))+cos(T_{new},v(j)))$ where $v(j) \in hard\_neg$
    \STATE Fine-tune BERT to minimize $loss$ and align $T_{new}$ and corresponding label representations from $O_{emb}$
 \\ \textbf{Testing Phase}
 \STATE Compute embeddings for test set $S$ using fine-tuned BERT $S_{emb} \gets BERT(S)$
 \STATE  $RO \gets sorted(Sim(S_{emb},O_{emb}))$, gives ranked set of labels
 \RETURN Top-k labels from $RO$
 \end{algorithmic} 
 \end{algorithm}

\subsection{Interactive cross-attention module for taxonomy aware input embeddings}
The primary goal of the proposed method is to align the vector representations of the input and the relevant hierarchical taxonomy labels. However, their vector representation sub-spaces may not be close to each other. This renders the alignment of the sub-spaces of the input representations and the corresponding label representations a hard problem. However, the tokens in the labels are related to terms in the content. The alignment approach would benefit from capturing the semantic relatedness between various hierarchical labels and a given content. The final vector representations are computed by fusing the information from hierarchical labels and the input. This would result in representations that are aware of the taxonomy and hence could help in performance improvement of the alignment task.
To achieve this, the proposed method first retrieves the label embedding closest to the input representation to capture the interaction between the labels and content which are related to each other (step 4 in Algorithm 1). This will reduce the noise induced by unrelated labels when fusing information from the input and the labels to compute \textit{taxonomy aware} input representations. 
           \[ T _{emb} = f_\theta(D_i)\]
           \[O_{emb}\gets g_\theta(O)\]
           \[O_{emb}^i \gets top_1(cos(T_{emb}, O_{emb}))\]
where $f_\theta$ is BERT and $g_\theta$ refers to sentence representation methods like Sentence BERT or Universal Sentence Encoder (USE). As this process occurs in the training loop, the closest label selected depends on the model parameters. The selection process is thereby dynamic and improves with updates to model parameters. When the model gets better at aligning the input and label representations, it will sample better top-1 hierarchical label closer to the input. Our task is modeled as a dense retrieval approach thereby, our goal is to improve the quality of the first label retrieved.

The step-5 in Algorithm 1 retrieves the top-k labels closest to the selected label. We do not compute attention with respect to all labels in the label space to obviate interference from unrelated labels and reduce the computational complexity of attention. We sample a small set of labels that are related to the label closest to the content representations at a given time in the training loop. This step helps to cluster similar labels which are closer to the input and to each other in the vector space. Also, it reduces the complexity involved in computing attention between the input and label representations. The labels sampled approximate the attention distribution well as shown below, adapting the property of attention from the work \cite{fast_transformers}.

The difference between the attention of inputs (Keys) and labels (Queries) can be bounded by the euclidean distance between the labels.

\textbf{Statement:} Given two label representations $O_{emb}^i, O_{emb}^j$, the difference between attention can be bounded as by the euclidean distance between the label representations.

To arrive at this result, we first start with the principle of Lipschitz continuity for Softmax. Given two queries $Q_i$ and $Q_j$,

\[||Softmax(Q_iK^T) - Softmax(Q_jK^T)||_2)\]
\[\leq \epsilon ||Q_iK^T - Q_jK^T||_2\] 
Let $\phi$ denote the Softmax operation for the rest of the section.

The Softmax operation has a Lipschitz constant less than 1\cite{softmax} which implies $\epsilon$ equals 1.
Hence the attention approximation is bounded by the euclidean distance between the queries\\
\begin{equation}
    ||\phi(Q_iK^T) - \phi(Q_jK^T)||_2\leq ||Q_i - Q_j||_2 ||K||_2
\end{equation}

Since $Q_i \gets W^Q * O_{emb}^i $ we can write the above equation as:
\begin{equation}
||\phi(Q_iK^T) - \phi(Q_jK^T)||_2\\
\leq ||O_{emb}^iW^Q - O_{emb}^jW^Q||_2 ||K||_2
\end{equation}
Since the norm of the weight matrix $W^Q$ is the largest eigenvalue of $((W^Q)^{T}W^Q)^{1/2}$ we modify the above equation as
\begin{equation}
||\phi(Q_iK^T) - \phi(Q_jK^T)||_2) \\
\leq ||O_{emb}^i - O_{emb}^j||_2 ||W^Q||_2 ||K||_2
\end{equation}
By equation 3, the difference between the attention can be hence bound as the euclidean distance between the label representations.

Following this bound, step-5 of Algorithm 1 samples top-k hierarchical labels ($L_r$) based on their proximity to the top-1 label in the representation space.

% We also provide an error bound for this approximation of label space by adapting the proof in \cite{fast_transformers} for our setting in Section \ref{proof}. We present the proof of error bound for the approximation for completion.

The input representations are projected to (K)ey and (V)alue matrices.
\[ K \gets W^K * T_{emb} ;  \ \  V \gets W^V * T_{emb} \]

where $W^K$ and $W^V$ are learnable weights.
The sampled label representations are projected to a (Q)uery  matrix
\[Q \gets W^Q * L_r\] and $W^Q$ is also learnable.

Then we propose an interactive (cross) attention mechanism where the compatibility between the labels (Q) and the inputs (K) are captured in form of an Attention matrix (A). Then the input representations are weighted by the attention weights in A to promote useful dimensions and drown out irrelevant ones.
\[        \alpha  =  \dfrac{Softmax(Q \cdot K^T)}{\sqrt{d_k}}
        % Softmax(x_i) & = & \dfrac{exp(x_i)}{\sum_j^N exp(x_j)} 
\]
\[        T_{new} \gets  \alpha \cdot V
\]
where, $T_{new}$ is now the vector representation that fuses the information from content and the representations of the sampled hierarchical labels (Step 8). In the above equation, $Q \in R^{l \times d}$, $K \in R^{n \times d}$ and $V \in R^{n \times d}$. Here $l$ is the number of labels sampled and n is the number of words in the questions (input content). Finally our output embedding from the interactive attention layer $T_{new} \in R^{l \times d}$ as there are $l$ labels sampled during training.

We finally average across the label dimension to compute a fixed length representation for the given question yielding

\[T_{new} = mean(T_{new},dim=0)\]

The computed representations are aligned with the corresponding label representations and pulled apart from representations of negative labels using a hinge rank loss function as explained in Section \ref{negatives}.
% \subsection{Theoretical guarantees for attention approximation}
% \label{proof}
% In the previous section, we described in detail our interactive attention mechanism for fusing information from the input (questions) and label (hierarchical taxonomy) spaces. However, we do not compute attention with respect to all labels in the label space. We sample a small set of labels that are related to the label closest to the content representations at a given time in the training loop. We now present an error bound for the approximation step of retrieving closest (Q)ueries (hierarchical labels) for computing the interactive attention. Adapting the proof in the work\cite{fast_transformers}, we provide error bounds for approximation error of the attention values computation.

% The difference between attention of inputs (Keys) and labels (Queries) can be bounded by the euclidean distance between the Queries.

% \textbf{Statement:} Given two Queries $Q_i, Q_j$ 
% \[||Softmax(Q_iK^T) - Softmax(Q_jK^T)||_2)\]
% \[\leq \epsilon ||Q_iK^T - Q_jK^T||_2\] \\
% The Softmax operation has Lipschitz constant less than 1\cite{softmax} which implies $\epsilon$ equals 1.
% Hence the attention approximation is bounded by the euclidean distance between the queries

% \[||Softmax(Q_iK^T) - Softmax(Q_jK^T)||_2) \]
% \[\leq ||Q_i - Q_j||_2 ||K||_2\] \\

   \begin{table*}
  \small
 \centering
\caption{Some samples from the QC-Science dataset}\label{tab1}
\begin{tabular}{|p{3.2cm}|p{4cm}|p{5.6cm}|}
\hline
Question &  Answer & Taxonomy\\
\hline
The value of electron gain enthalpy of chlorine is more than that of fluorine. Give reasons &  Fluorine atom is small  so electron charge density on F atom is very high & Science$\xrightarrow{}$chemistry$\xrightarrow{}$classification of elements and periodicity in properties\\ \hline
What are artificial sweetening agents? & The chemical substances which are sweet in taste but do not add any calorie & Science$\xrightarrow{}$chemistry$\xrightarrow{}$chemistry in everyday life\\
\hline
\end{tabular}
\end{table*}

             \begin{table*}
       \small
 \centering
\caption{Comparison of different representation methods for hierarchical labels}\label{tab1}
\begin{tabular}{p{3.7cm}|p{4.8cm}|p{4.8cm}|c}
 Method & Label1 (L1)& Label2 (L2) & cos(L1, L2)\\
\hline\hline
Sentence-BERT &  science $\xrightarrow{}$ physics $\xrightarrow{}$ electricity &  science $\xrightarrow{}$ chemistry $\xrightarrow{}$ acids & 0.3072\\ 
Sent2vec & science $\xrightarrow{}$ physics $\xrightarrow{}$ electricity &  science $\xrightarrow{}$ chemistry $\xrightarrow{}$ acids & 0.6242\\ 
GloVe & science $\xrightarrow{}$ physics $\xrightarrow{}$ electricity &  science $\xrightarrow{}$ chemistry $\xrightarrow{}$ acids & 0.6632\\ 

\hline

\end{tabular}
\end{table*}

\subsection{Adaptive hard-negative sampling}
\label{negatives}
After the modified input representations are obtained, we proceed to the training step, where the input representations are aligned with the corresponding label representations and pushed apart from the representations of negative labels using a hinge rank loss \cite{frome2013devise}.

For learning representations that disentangle the vector representations of positive and negative labels, we sample hard negatives when optimizing the loss function. The hard negatives are those hierarchical labels with a high semantic relatedness score (cosine similarity) to the input questions but are not the correct hierarchical labels. We sample them using the following equation:

\[ hard\_neg \gets top\_k(cos(T_{new},Index(labels))) \]

where $Index(labels)$ refers to the in-batch hierarchical labels and $label \notin hard\_neg$, $k<batch\_size$. We experiment with different values of k and observe that k =5 gives the best results. After sampling the ground truth hierarchical label as positive and the hard negatives, the hinge rank loss is employed to optimize for the alignment of input and label representations.

The hinge ranking loss is defined as :
\[         L(text,l)\gets \sum_{j \neq l}max(0,\delta-cos(T_{emb},v(l)) \\ +cos(T_{emb},v(j)))
 \]
 \[L(text,l) \gets \dfrac{L(text,l)}{len(hard\_neg)}\]
 where, $j \in hard\_neg$, $T_{emb}$ denotes the input text embeddings from BERT,
$v(label)$ denotes the vector representation of the correct label, $v(j)$ denotes the vector representation of an incorrect label. The margin value was set to 0.1, which is a fraction of the norm of the embedding vectors (1.0), and resulted in the best performance.

The hard negatives are sampled dynamically during the training and hence are a function of model parameters.

\[hard\_neg \gets f(\theta)\]
where $f(\theta)$ denotes the BERT model and $\theta$ denotes the model parameters.

At each iteration in the training loop, we sample the incorrect labels which are closer in the vector space to the input representations. This ensures that the hard negatives improve as the model parameters are updated to better align with the correct label representations. We conduct several ablation studies to compare with random negative sampling and demonstrate that the proposed method aids in high recall retrieval.

 \section{Experiments}
 \label{experiments}
 In this section, we discuss the baselines, experimental setup and the datasets used. All experiments are carried out on Google colab.

 \subsection{Datasets}
 To evaluate the efficacy of the method \textit{TagRec++}, we perform experiments on the following datasets:
 \let\labelitemi\labelitemii
 \begin{itemize}
     \item \textbf{QC-Science}: This dataset contains 47832 question-answer pairs belonging to the science domain tagged with hierarchical labels of the form subject - chapter - topic. The dataset was collected with assistance from a leading e-learning platform. The dataset consists of 40895 training samples, 2153 validation samples and 4784 test samples. Some samples are shown in Table 1. The average number of words per question is 37.14, and per answer, it is 32.01.
     \item \textbf{ARC} \cite{xumulti}: This dataset consists of 7775 science multiple choice exam questions with answer options and 406 hierarchical labels. The average number of words per question is 20.5. The number of samples in the train, validation, and test sets are 5597, 778 and 1400, respectively.

    %  \item \textbf{Learning Objectives}: This dataset consists of 417 learning objectives collected from the \textit{"What you learnt"} section from grades 8 to 10 science textbooks (K-12 system). The corresponding learning taxonomy was derived from the "Table of contents" (TOC) of the textbooks. 

 \end{itemize}

In our experiments, the question and the answer are concatenated and used as the input to the model (BERT), and the hierarchical taxonomy is considered as the label. The number of tokens of each question-answer pair is within 512 and hence can be accommodated within the context limit of BERT.

\subsection{Analysis of sentence representation methods for representing the hierarchical labels}
In this section, we provide an analysis of results from a meta-experiment to decide the best sentence representation methods for the hierarchical labels (learning taxonomy). We embed the hierarchical labels using methods like Sent2Vec \cite{pagliardini2017unsupervised} and Sentence-BERT \cite{reimers-gurevych-2019-sentence}.
Additionally, we also leverage word embedding methods like GloVe to represent the hierarchical labels by averaging the representations of the tokens in the hierarchical label. We then compute the semantic similarity between the resulting representations of two different hierarchical labels, as shown in Table 2. From Table 2, we observe that though "science $\xrightarrow{}$ physics $\xrightarrow{}$ electricity" and "science $\xrightarrow{}$ chemistry $\xrightarrow{}$ acids" are different, a high similarity score is obtained between representations obtained using GloVe embeddings. This may be due to the observation that averaging word vectors can result in loss of information. Additionally, the context of words like physics is not taken into account when encoding the word electricity. Additionally, the words "physics" and "chemistry" are co-hyponyms which may result in their vectors being close in the continuous vector space when using traditional static embedding methods. We also observe that static sentence embeddings from Sent2Vec are also unable to capture the context of the tokens in the labels as the representations obtained from Sent2Vec result in a  high similarity score. However, we observe that the representations obtained using sentence transformer-based methods like Sentence-BERT are not very similar, as indicated by the similarity score. This indicates that Sentence-BERT is able to produce meaningful sentence representations leveraging the context of tokens for the hierarchical labels. We also observe that Sentence-BERT outputs high similarity scores for semantically related hierarchical labels. We provide a detailed comparison of methods using different vector representation methods in Section 5 as this analysis is not exhaustive.

       \begin{table*}
  \small
 \centering \caption{Performance comparison of TagRec++ with variants and baselines, $\dagger$ indicates TagRec++'s significant improvement at 0.001 level using \textit{t-test}}\label{results1}
 \fontsize{9.5}{9}\selectfont
\centering
\begin{tabular}{p{2.1cm}|p{7.2cm}|p{1.2cm}|p{1.2cm}|p{1.2cm}}
Dataset & Method &  R@1 & R@3& R@5\\
\hline\hline
 & TagRec(BERT+USE) &  0.54 &0.78 & 0.86 \\ 
&TagRec(BERT+SB) & 0.53 &  0.77 &  0.85 \\
 & TagRec++(BERT+USE) (ours) & \bf 0.65 $\dagger$ & \bf 0.84 $\dagger$ & \bf0.89 $\dagger$ \\ 
&TagRec++(BERT+SB) (ours) & \bf 0.65 $\dagger$ & \bf 0.85 $\dagger$ & \bf 0.90 $\dagger$ \\ \cline{2-5}
QC-Science&BERT+sent2vec & 0.43 &0.70 & 0.79\\ 
&Twin BERT \cite{twinbert} & 0.32 &0.60 & 0.72\\ 
&BERT+GloVe & 0.39 &0.65 & 0.76\\ \cline{2-5}
&BERT classification (label relation) \cite{xumulti} \ & 0.19 &0.33 & 0.39 \\ 
&BERT classification (prototypes) \cite{snell2017prototypical} & 0.54 &0.75 & 0.83 \\ \cline{2-5}
& Pretrained Sent\_BERT  & 0.11 &0.22 & 0.30 \\ 

\hline
 & TagRec(BERT+USE) &  0.35 &0.55 & 0.67 \\ 
&TagRec(BERT+SB) & 0.36 & 0.55 &  0.65 \\
& TagRec++(BERT+USE) (ours) & \bf 0.48 $\dagger$ & \bf 0.66 $\dagger$ & \bf0.75 $\dagger$ \\ 
&TagRec++(BERT+SB) (ours) & \bf 0.49 $\dagger$ & \bf 0.71 $\dagger$ & \bf 0.78 $\dagger$ \\\cline{2-5} 
ARC &BERT+sent2vec & 0.22 &0.43 & 0.55\\ 
&Twin BERT \cite{twinbert} & 0.14 &0.31 & 0.46\\ 
&BERT+GloVe & 0.23 &0.43 & 0.56\\ \cline{2-5}
&BERT classification (label relation) \cite{xumulti} \ & 0.11 &0.21 & 0.27 \\ 
&BERT classification (prototypes) \cite{snell2017prototypical} & 0.35 &0.54 & 0.64 \\ \cline{2-5}
& Pretrained Sent\_BERT  & 0.12 &0.24 & 0.31 \\ 
\hline
\end{tabular}
\end{table*}
  \begin{table*}
% \setlength{\textfloatsep}{0.1cm}
% \vspace{-1mm}
\caption{Ablation analysis of TagRec++}
 \fontsize{9.5}{9}\selectfont
\centering
\label{ablation:results}
\begin{tabular}{p{2.1cm}|p{6.4cm}|p{1cm}|p{1cm}|p{1cm}}
Dataset & Method &  R@1 & R@3& R@5\\
\hline\hline
 & 
  TagRec++(BERT+USE) (proposed method) & \bf 0.65 & \bf 0.84 & \bf0.89 \\ 
&TagRec++(BERT+SB) (proposed method) & \bf 0.65 & \bf 0.85 & \bf 0.90 \\
QC-Science & TagRec++(BERT+USE) (- attention) &  0.62 &  0.83 & 0.88 \\ 
 &TagRec++(BERT+SB) (- attention) &  0.62 &  0.83 &  0.88 \\
 & TagRec++(BERT+USE) (- hard-negatives) & 0.57 & 0.81 & 0.86 \\ 
&TagRec++(BERT+SB) (- hard-negatives) &  0.56 & 0.80 & 0.87\\
\hline

& TagRec++(BERT+USE) (proposed method) & \bf 0.47 & \bf 0.69 & \bf0.75 \\ 
&TagRec++(BERT+SB) (proposed method) & \bf 0.49 & \bf 0.71 & \bf 0.77 \\
ARC  & TagRec++(BERT+USE) (- attention) & 0.41 & 0.61 & 0.70 \\ 
&TagRec++(BERT+SB) (- attention) &  0.44 & 0.64 & 0.74 \\ &  TagRec++(BERT+USE) (- hard-negatives) & 0.39 & 0.60 & 0.72 \\ 
&TagRec++(BERT+SB) (- hard-negatives) &  0.45 & 0.66 & 0.74 \\\cline{2-5}
\hline
\end{tabular}
\end{table*}

\subsection{Methods and Experimental setup}
 We compare TagRec with flat multi-class classification methods and other state-of-the-art methods. In TagRec, the labels are represented using transformer based sentence representation methods like Sentence-BERT (Sent\_BERT) \cite{reimers-gurevych-2019-sentence} or Universal Sentence Encoder \cite{cer-etal-2018-universal}.
 
The methods we compare against are:
\let\labelitemi\labelitemii
\begin{itemize}
  \setlength\itemsep{0.5em}

    \item \textbf{BERT+Sent2Vec} : In this method, the training and testing phases are similar to TagRec. The label representations are obtained using Sent2vec \cite{pagliardini2017unsupervised} instead of USE or Sent\_BERT.
    \item \textbf{BERT+GloVE} : In this method, the labels are represented as the average of the word embeddings of their constituent words. The word embeddings are obtained from GloVe.
    \setlength{\parindent}{0pt}
    \[V(label) = mean((Gl(subject),Gl(chapter),Gl(topic)))\]
    where $V(label)$ denotes vector representation of the label, $Gl$ denotes GloVe pre-trained model. The training and testing phases are the same as TagRec.

    \item \textbf{Twin BERT}:  This method is reproduced from the work Twin BERT \cite{twinbert}. In this method, a pre-trained BERT model is fine-tuned to represent the labels in a continuous vector space. The label representations correspond to the first token of the last layer hidden state denoted as [CLS] in BERT. The two BERT models in the two-tower architecture are fine-tuned simultaneously. 

    \item \textbf{BERT multi-class}  (label relation)  \cite{xumulti}: In this method,  the hierarchical labels are flattened and encoded, resulting in a multi-class classification method. Then we fine-tune a pre-trained BERT model for categorizing the input content to the labels. During inference, the representations for the inputs and labels are computed using the fine-tuned model. Then the labels are retrieved and ranked according to the cosine similarity scores computed between the input text representations and the label representations.

    \item \textbf{BERT multi-class} (prototypes) \cite{snell2017prototypical}: To provide a fair comparison with \textit{TagRec++}, we propose another baseline that considers the inter-sample similarity than similarity between inputs and labels. A BERT model is fine-tuned like the previous baseline. Then for each class, we compute the mean of the embeddings of random samples from the training set to serve as the prototype for the class. The vector representation for each selected sample is obtained by the concatenation of the [CLS] token obtained from the last 4 layers of the fine-tuned model. This method of vector representation gives the best performance. After obtaining the class prototypes, at inference, we obtain the embeddings for each test sample and compute cosine similarity with the class prototype embeddings. Then the top-k classes ranked as per cosine similarity are returned.
    
    \item \textbf{Pretrained Sent\_BERT}: We implement a baseline where the input texts and labels are encoded using a pre-trained Sentence-BERT model. For each input, the top-k labels are retrieved by computing cosine similarity between the input and the label representations.
    
    \item \textbf{TagRec} \cite{Tagrec}: This can be considered as ablation or preliminary version of TagRec++, where the attention mechanism and adaptive hard negative sampling are the missing modules when compared to TagRec++.
        \end{itemize}
        
Some other ablations we perform are:
\begin{itemize}
    \item \textbf{TagRec++ (-attention)}: We perform an experiment with the removal of the interactive attention mechanism explained in Section \ref{methods} from TagRec++ and compare it with the original approach to test the effectiveness of the proposed attention mechanism. We perform this experiment for both variants of TagRec++.
        \item \textbf{TagRec++ (-hard-negatives)}: In this variant of TagRec++, instead of the adaptive in-batch hard negative sampling, we replace them with adaptive random negatives. We sample the same number of negatives for a fair comparison. We perform this experiment to analyze the impact of our adaptive hard-negative sampling approach on the final performance of the proposed approach.
\end{itemize}
 
\section{Results and Analysis}  
\label{results}
The performance comparison of TagRec++ with baselines and other variants can be observed from Table \ref{results1}. We observe that TagRec++ outperforms all approaches as measured by Recall@k (R@k). \\

\begin{table*}[h!]
\small
\centering
\caption{Examples demonstrating the performance for unseen labels at test time.}
\label{unseen:labels}
\begin{tabular}{p{4.2cm}|p{2.9cm}|p{4cm}|p{2.5cm}}
% \hline
% \small
Question text & Ground truth  & Top 2 predictions & Method\\ \hline \hline
A boy can see his face when he looks into a calm pond. Which physical property of the pond makes this happen? (A) flexibility (B) reflectiveness (C) temperature (D) volume& \multirow{5}{3cm}{matter$\xrightarrow{}$properties  of material$\xrightarrow{}$reflect} & \multirow{5}{3.2cm} {matter$\xrightarrow{}$properties of material$\xrightarrow{}$flex and}  \multirow{9}{3cm}{\textbf{matter$\xrightarrow{}$properties of material$\xrightarrow{}$reflect}} & \multirow{7}{2cm}{TagRec++ (BERT+USE) }
 %\cline{1-1} &
\\\cline{3-4}
% A: Fluorine atom is small, so electron charge density on F atom is very high and concentrated on very small area & & \\ \hline

 &  & matter$\xrightarrow{}$properties of objects$\xrightarrow{}$mass  \\ & & and \\& & matter$\xrightarrow{}$properties of objects$\xrightarrow{}$density & Twin BERT \cite{twinbert} \\\cline{3-4}
  &  & matter$\xrightarrow{}$states$\xrightarrow{}$solid and \\ & & matter$\xrightarrow{}$properties of material$\xrightarrow{}$density & BERT+GloVe \\\cline{3-4}
&  &   matter$\xrightarrow{}$properties of material$\xrightarrow{}$specific heat and \\ & & matter$\xrightarrow{}$properties \ of material& BERT+sent2vec \\ \hline
Which object best reflects light? (A) gray door (B) white floor (C) black sweater (D) brown carpet & \multirow{1}{3cm}{matter$\xrightarrow{}$ properties  of material$\xrightarrow{}$reflect} & \multirow{1}{3cm} {energy$\xrightarrow{}$light$\xrightarrow{}$reflect and}  \multirow{4}{3.2cm}{\textbf{matter$\xrightarrow{}$properties of material$\xrightarrow{}$reflect}} & \multirow{5}{2cm}{TagRec++ (BERT+USE)} 
\\ \cline{3-4}

 &  & energy$\xrightarrow{}$thermal$\xrightarrow{}$\\ & & radiation and \\ & & energy$\xrightarrow{}$light$\xrightarrow{}$generic properties & Twin BERT \cite{twinbert} \\\cline{3-4}
 &  & energy$\xrightarrow{}$light and \\ & & energy$\xrightarrow{}$light$\xrightarrow{}$refract & BERT+GloVe \\\cline{3-4}
&  &   energy$\xrightarrow{}$light$\xrightarrow{}$reflect and \\ & & energy$\xrightarrow{}$light$\xrightarrow{}$refract& BERT+sent2vec \\ \hline
% A: The chemical substances which are sweet in taste but do not add any calorie & &\\
\end{tabular}
\end{table*}

\textbf{Advantage of contextualized sentence representations}: \\
We observe that contextualized embedding methods for labels provide the best performance. This is evident from the table as TagRec++(BERT+USE) and TagRec++(BERT+Sent\_BERT) outperform approaches which leverage static sentence embedding methods like BERT+Sent2Vec and BERT+GloVe. This is because transformer-based encoding methods use self-attention to produce better representations. In addition, the Sentence-BERT and the Universal Sentence Encoder models are ideal for retrieval based tasks as they were pre-trained on semantic text similarity (STS) tasks. \\

\textbf{Performance comparison with flat classification methods}:
Finally, we observe that the TagRec++ method outperforms the flat multi-class classification based baselines, confirming the hypothesis that capturing the semantic relatedness between the terms in the input and tokens in the hierarchical labels results in better representations. This is pivotal to the question-answer pair categorization task as the technical terms in the short input text are semantically related to the tokens in the label. The baseline (BERT label relation) performs poorly as it has not been explicitly trained to align the input and the hierarchical label representations. The representations obtained through the flat multi-class classification approach have no notion of semantic relatedness between the content and label representations. But the prototypical embeddings baseline performs better as the classification is done based on semantic matching between train and test sample representations. However, this baseline also has no notion of semantic relatedness between the input and label representations. Hence it does not perform well when compared to our proposed method, TagRec++. Moreover, this baseline cannot also adapt to changes in the label space and requires a change in the final classification layer and \textit{retraining}. We also observe that the baseline of semantic matching using pre-trained sentence BERT does not work well.\\

\textbf{Are the results statistically significant?}
To confirm the efficacy of \textit{TagRec++}, we perform statistical significance tests and observe that the predicted results are statistically significant over \textit{TagRec}. For instance, for Recall@5 we observe that the predicted outputs from TagRec++ are statistically significant (\textit{t-test}) with p-values \textbf{0.0000499} and \textbf{0.0000244} for \textit{QC-Science} and \textit{ARC} respectively.
\\

 \begin{table}
 \label{zeroshot:lo}
 \small
 \centering
\caption{Performance comparison for zero-shot learning objective categorization}\label{tab1}
\begin{tabular}{p{5.6cm}|p{1cm}|p{1cm}}
 Method & R@1& R@2\\
\hline\hline
TagRec++(BERT+SB) (ours) & \bf 0.82 & \bf 0.94 \\ 
 TagRec++(BERT+USE) (ours) &  0.79 &0.93\\ 
 TagRec(BERT+USE)  &  0.69 &0.85\\ 
TagRec(BERT+SB)  &  0.77 & 0.91 \\ 
BERT+sent2vec & 0.49 &0.64\\ 
Twin BERT \cite{twinbert} & 0.54 &0.79\\ 
BERT+GloVe & 0.62 &0.84\\ 
BERT classification (label relation) \cite{xumulti} \ & 0.46 &0.59\\ 
BERT classification (prototypes) \cite{snell2017prototypical} & 0.60 &0.76 \\ 
 Pretrained Sent\_BERT  & 0.39 &0.54  \\ 
\hline

\end{tabular}
\end{table}

 \begin{table}
 \small
 \centering
\caption{Ablation results for zero-shot evaluation on learning objective categorization}\label{zeroshot:ablation}
\begin{tabular}{p{5.6cm}|p{1cm}|p{1cm}}
 Method & R@1& R@2\\
\hline\hline
TagRec++(BERT+SB) (ours) & \bf 0.82 &  0.94 \\ 
TagRec++(BERT+SB) (-attention) &  0.80 &  0.94 \\ 
 TagRec++(BERT+SB) (-hard-negatives) &  0.80 &0.93\\ 
\hline
\end{tabular}
\end{table}

\textbf{Ablation studies:}
We also perform several ablation analyses of the proposed TagRec++ approach. As observed in Table \ref{ablation:results} we compare TagRec++ with and without the proposed interactive attention mechanism. We see a clear performance difference confirming the hypothesis that the interactive attention mechanism is crucial for high recall retrieval as it captures the relatedness between the tokens in the hierarchical labels and the terms in the input content (questions).

We also performed another ablation study to ascertain the effectiveness of the proposed in-batch hard negative sampling. Instead of sampling in-batch hard negatives, we sample random negatives. The random negatives are also sampled dynamically for a fair comparison. From Table \ref{ablation:results}, we can observe that the proposed in-batch hard negative sampling works better than the random negative sampling. In the hard negative sampling approach, as the model is trained to align the content and appropriate label representations, it gets better at sampling hard negatives which result in high-recall retrieval. 
\\
 \begin{table*}
 \small
 \centering
\caption{Performance comparison for each epoch on ARC dataset: TagRec++ vs TagRec++ (-hard-negatives}\label{epochwise}
\begin{tabular}{p{5.5cm}|c|c|c|c|c}
&  \multicolumn{5}{c}{\bf Epochs}\\\cline{2-6} 
 Method&  \bf2 &4 &6 &8 &10 \\
\hline\hline
TagRec++(BERT+SB) (ours) & \bf 0.30 
&\bf 0.40 
& \bf 0.43
& \bf 0.45  & \bf 0.45 \\
\hline
TagRec++(BERT+SB) (-hard-negatives) & 0.22 
& 0.33 
& 0.35
& 0.39
 & 0.39 \\
\hline
\end{tabular}
\end{table*}

 \begin{table*}
 \small
 \centering
\caption{Qualitative analysis of top-3 hard-negatives sampled on QC-Science dataset}\label{hardnegepoch}
\begin{tabular}{p{4.8cm}|c|c|p{2.8cm}|p{2.8cm}|p{3cm}}
&  & & \multicolumn{3}{c}{\bf Hard negatives}\\\cline{4-6} 
Question & Answer & Epoch &  \bf1 & 2 &3 \\
\hline\hline
& & 
 1 & science $\xrightarrow[]{}$ physics $\xrightarrow{}$ work, energy and power &physics $\xrightarrow{}$  part - ii $\xrightarrow{}$ mechanical properties of fluids & physics $\xrightarrow{}$  part - ii $\xrightarrow{}$ thermal properties of matter\\ \cline{3-6}
 In the given transistor circuit, the base current is 35 $\mu$A. The value of R b is & 200 Omega & 5 & science $\xrightarrow{}$  physics $\xrightarrow{}$ communication systems &physics $\xrightarrow{}$  part - i $\xrightarrow{}$ magnetism and matter & physics $\xrightarrow{}$  part - i $\xrightarrow{}$ system of particles and rotational motion\\ \cline{3-6}
 & & 10 & physics $\xrightarrow{}$  part i $\xrightarrow{}$ magnetic effects of current &physics $\xrightarrow{}$  part - i $\xrightarrow{}$ magnetism and matter  & physics $\xrightarrow{}$  part - i $\xrightarrow{}$ moving charges and magnetism\\
\hline
\end{tabular}
\end{table*}
\textbf{Performance on unseen labels:}
The TagRec++ was also able to adapt to changes in the label space. For instance, in the \textit{ARC} dataset, two samples in the test set were tagged with \textit{"matter$\xrightarrow{}$properties of material$\xrightarrow{}$reflect"} unseen during the training phase as shown in Table \ref{unseen:labels}. At test time, the label \textit{"matter$\xrightarrow{}$properties of material$\xrightarrow{}$reflect"} appeared in top 2 predictions output by the proposed method (TagRec++ (BERT + USE)) for the two samples. We also observe that for the method (TagRec++ (BERT + Sent\_BERT)) the label \textit{"matter$\xrightarrow{}$properties of material$\xrightarrow{}$reflect"} appears in its top 5 predictions. We observe that for other baselines shown in Table \ref{unseen:labels} the correct label does not get retrieved even in top-10 results. The top 2 results retrieved from other methods for the samples are shown in Table \ref{unseen:labels}. Similar results can be observed for the BERT classification (label relation) and BERT classification (prototypes) baselines. We do not show them in Table \ref{unseen:labels} owing to space constraints. The top 2 predictions from BERT classification (prototypes) baseline for example 1 in Table \ref{unseen:labels} are \textit{matter$\xrightarrow{}$properties of objects$\xrightarrow{}$temperature} and \textit{matter$\xrightarrow{}$properties of objects$\xrightarrow{}$shape}.

 \begin{table}
 \small
 \centering
\caption{Performance comparison for different values of k when sampling top-k taxonomy tags for attention in TagRec++ }\label{attentiontopk}
\begin{tabular}{p{3cm}|p{1cm}|p{1cm}|p{1cm}}
 \# of tags for attention & R@1 & R@3& R@5\\
\hline\hline
1 &  0.61 & 0.82 & 0.88 \\ 
5 & \bf 0.65 & \bf 0.85 & \bf 0.90 \\
10 & 0.63 &  0.85 & 0.89 \\

\hline
\end{tabular}
\end{table}

 \begin{table*}
 \small
 \centering
\caption{Qualitative analysis of top-3 tags sampled on QC-Science dataset for the cross-attention mechanism}\label{attentionepoch}
\begin{tabular}{p{4.5cm}|c|c|p{3cm}|p{3cm}|p{3cm}}
&  & & \multicolumn{3}{c}{\bf hierarchical tags for attention}\\\cline{4-6} 
Question & Answer & Epoch &  \bf1 & 2 &3 \\
\hline\hline
& & 
 1 & physics $\xrightarrow{}$  part - i $\xrightarrow{}$ magnetism and matter & physics $\xrightarrow{}$  part - i $\xrightarrow{}$ moving charges and magnetism & science $\xrightarrow{}$physics$\xrightarrow{}$magnetic effects of electric current\\\cline{3-6}
 In the given transistor circuit, the base current is 35 $\mu$A. The value of R b is & 200 Omega & 5 & physics $\xrightarrow{}$  part - i $\xrightarrow{}$ electromagnetic waves &physics $\xrightarrow{}$  part - i $\xrightarrow{}$ alternating current & physics $\xrightarrow{}$  part - i $\xrightarrow{}$ electrostatic potential and capacitance\\ \cline{3-6}
 & & 10 & physics $\xrightarrow{}$  part - i $\xrightarrow{}$ alternating current &physics $\xrightarrow{}$  part - i $\xrightarrow{}$ current electricity & physics $\xrightarrow{}$  part - i $\xrightarrow{}$ electric current and it's effects\\
\hline
\end{tabular}
\end{table*}
For example 2, in Table \ref{unseen:labels}, the top 2 predictions from BERT classification (prototypes) are \textit{energy$\xrightarrow{}$light$\xrightarrow{}$reflect} and \textit{matter$\xrightarrow{}$properties of material$\xrightarrow{}$color}.

The top 2 predictions from BERT classification (label relation) baseline for example 1 in Table \ref{unseen:labels} are \textit{matter$\xrightarrow{}$properties of objects$\xrightarrow{}$ density} and \textit{matter$\xrightarrow{}$\\properties of material$\xrightarrow{}$density}. For example 2, in Table \ref{unseen:labels}, the top 2 predictions from BERT classification (label relation) are \textit{energy$\xrightarrow{}$light$\xrightarrow{}$refract} and \textit{matter$\xrightarrow{}$properties of material$\xrightarrow{}$luster}. This validates our hypothesis that the proposed method can adapt to new labels without changes in model architecture and retraining, unlike existing methods.

\textbf{Zero-shot performance:} 
We curated a set of learning objectives from K-12 textbooks to test the ability of \textit{TagRec++} to tag related short learning content without training. This experiment is performed to observe the zero-shot abilities of \textit{TagRec++}. We observe that \textit{TagRec++} outperforms existing approaches as measured by Recall@k as shown in Table \ref{zeroshot:lo}. This demonstrates that the proposed approach also leads to high recall retrieval in a zero-shot setting.

We also perform certain ablation studies for the zero-shot setting by removing the interactive attention component and the in-batch hard negatives sampling approach as shown in Table \ref{zeroshot:ablation}. We observe that \textit{TagRec++} achieves the highest performance indicating the significance of the proposed attention mechanism and hard negative sampling method.
\\

\textbf{Epochwise comparison of hard negatives sampling vs random negatives sampling approaches:}

The in-batch dynamic negative sampling is based on the hypothesis that the model improves with training and samples better hard negatives leading to high recall retrieval when compared to dynamic random negatives sampling. To test this hypothesis, apart from the ablation shown in Table \ref{ablation:results}, we also perform an epochwise comparison of dynamic hard negatives and dynamic random negatives as shown in Table \ref{epochwise}. We observe that \textit{TagRec++} with dynamic hard negatives has a higher recall in each epoch due to better sampling when compared to dynamic random negatives. This demonstrates that hard negatives lead to better recall than random negatives. Also, the proposed dynamic sampling approach leads to better hard negatives as the model learns to align the content and hierarchical label representation sub-spaces.

\textbf{Qualitative analysis of the hard-negatives sampled dynamically during training}

We analyze the hard-negatives sampled in the training loop to determine if the quality of hard negatives increases as training progresses. The observation for a sample is shown in Table \ref{hardnegepoch}. We observe in epoch one the hard negatives are centered around \textit{physics} subject but the topics are not related to the input. The ground truth label for the question shown in the table is centered around \textit{electrical circuits}. We observe that as training progresses, the top-2 labels are centered around \textit{magnetism}, \textit{communication systems} but do not correspond to the correct theme of \textit{electrical circuits}, rendering them as hard negatives. This demonstrates that dynamic sampling of in-batch hard-negatives improves with training. We observe a similar phenomenon for other samples too, which are not attached here due to space constraints.

\textbf{Analysis of top-k tags sampled for cross-attention module:}

We vary the value of k for the top-k tags (hierarchical labels) sampled for cross-attention discussed in Section 3.1. The results are shown in Table \ref{attentiontopk}. We observe that the highest R@k is achieved for value of 5. When only one tag is sampled it contains very less information as demonstrated by the values of R@k. We also observe sampling ten tags to fuse the taxonomy information with the input leads to a lower R@k. This maybe due to noise induced by the tags less related to the input. Hence we set k in top-k tags sampled to 5.

We also perform the qualitative analysis of tags sampled as shown in Table \ref{attentionepoch}. We observe that though in first epoch, we get tags related to \textit{magnetism} as the training progresses, in later epochs, we get most tags relevant to \textit{electricity} which are closer to the ground truth label \textit{semiconductors and electrical circuits}. This helps the model capture the information in tokens from various tags relevant to the topic and the terms in the input question. 

\section{Conclusion and Future Work}
\label{conclusion}
We proposed a novel approach TagRec++ for tagging content to hierarchical taxonomy. The proposed approach can be used in online systems to onboard and tag content to standardized taxonomy at scale. We proposed an adaptive in-batch hard-negative sampling approach for achieving high recall retrieval. We also propose a cross-attention approach where we fuse the information from the content and the hierarchical label embeddings to capture the interaction between the tokens in the hierarchical labels and the terms in the input content. We observe that the proposed approach outperforms \textit{TagRec}, other baselines and achieves high recall retrieval.

In the future, we plan to embed the hierarchical labels in the hyperbolic space as they are more suited to represent hierarchical data. This would also require embedding the input content representations in the hyperbolic space to capture the interaction between the content and label representations. We also plan to test the ability of our model to adapt to changes in the label space at scale by deploying the model in a K-12 system different from training datasets.
\section*{Acknowledgements}
The authors acknowledge the support of Extramarks Education India Pvt. Ltd., SERB, FICCI (PM fellowship), Infosys Centre for AI and TiH Anubhuti (IIITD).

\bibliographystyle{IEEEtran}
\bibliography{tagrecbib}

\begin{thebibliography}{10}
\providecommand{\url}[1]{#1}
\csname url@samestyle\endcsname
\providecommand{\newblock}{\relax}
\providecommand{\bibinfo}[2]{#2}
\providecommand{\BIBentrySTDinterwordspacing}{\spaceskip=0pt\relax}
\providecommand{\BIBentryALTinterwordstretchfactor}{4}
\providecommand{\BIBentryALTinterwordspacing}{\spaceskip=\fontdimen2\font plus
\BIBentryALTinterwordstretchfactor\fontdimen3\font minus
  \fontdimen4\font\relax}
\providecommand{\BIBforeignlanguage}[2]{{%
\expandafter\ifx\csname l@#1\endcsname\relax
\typeout{** WARNING: IEEEtran.bst: No hyphenation pattern has been}%
\typeout{** loaded for the language `#1'. Using the pattern for}%
\typeout{** the default language instead.}%
\else
\language=\csname l@#1\endcsname
\fi
#2}}
\providecommand{\BIBdecl}{\relax}
\BIBdecl

\bibitem{xumulti}
D.~Xu, P.~Jansen, J.~Martin, Z.~Xie, V.~Yadav, H.~Tayyar~Madabushi, O.~Tafjord,
  and P.~Clark, ``\BIBforeignlanguage{English}{Multi-class hierarchical
  question classification for multiple choice science exams},'' in
  \emph{\BIBforeignlanguage{English}{Proceedings of the 12th Language Resources
  and Evaluation Conference}}.\hskip 1em plus 0.5em minus 0.4em\relax
  Marseille, France: European Language Resources Association, May 2020, pp.
  5370--5382.

\bibitem{kozareva2015everyone}
Z.~Kozareva, ``Everyone likes shopping! multi-class product categorization for
  e-commerce,'' in \emph{Proceedings of the 2015 Conference of the North
  American Chapter of the Association for Computational Linguistics: Human
  Language Technologies}, 2015, pp. 1329--1333.

\bibitem{flat1}
\BIBentryALTinterwordspacing
Y.~Yang and X.~Liu, ``A re-examination of text categorization methods,'' in
  \emph{Proceedings of the 22nd Annual International ACM SIGIR Conference on
  Research and Development in Information Retrieval}, ser. SIGIR '99.\hskip 1em
  plus 0.5em minus 0.4em\relax New York, NY, USA: Association for Computing
  Machinery, 1999, p. 42–49. [Online]. Available:
  \url{https://doi.org/10.1145/312624.312647}
\BIBentrySTDinterwordspacing

\bibitem{flat2}
Y.~Yang and J.~O. Pedersen, ``A comparative study on feature selection in text
  categorization,'' in \emph{Proceedings of the Fourteenth International
  Conference on Machine Learning}, ser. ICML '97.\hskip 1em plus 0.5em minus
  0.4em\relax San Francisco, CA, USA: Morgan Kaufmann Publishers Inc., 1997, p.
  412–420.

\bibitem{flat3}
X.~Qiu, J.~Zhou, and X.~Huang, ``An effective feature selection method for text
  categorization,'' in \emph{Proceedings of the 15th Pacific-Asia Conference on
  Advances in Knowledge Discovery and Data Mining - Volume Part I}, ser.
  PAKDD'11.\hskip 1em plus 0.5em minus 0.4em\relax Berlin, Heidelberg:
  Springer-Verlag, 2011, p. 50–61.

\bibitem{twinbert}
W.~Lu, J.~Jiao, and R.~Zhang, ``Twinbert: Distilling knowledge to
  twin-structured compressed bert models for large-scale retrieval,'' ser. CIKM
  '20.\hskip 1em plus 0.5em minus 0.4em\relax New York, NY, USA: Association
  for Computing Machinery, 2020, p. 2645–2652.

\bibitem{Tagrec}
\BIBentryALTinterwordspacing
V.~Venktesh, M.~Mohania, and V.~Goyal, ``Tagrec: Automated tagging of questions
  with hierarchical learning taxonomy,'' 2021. [Online]. Available:
  \url{https://arxiv.org/abs/2107.10649}
\BIBentrySTDinterwordspacing

\bibitem{article}
W.~Xia, W.~Zhu, B.~Liao, M.~Chen, L.~Cai, and L.~Huang, ``Novel architecture
  for long short-term memory used in question classification,''
  \emph{Neurocomputing}, vol. 299, 03 2018.

\bibitem{yu2012product}
H.-F. Yu, C.-H. Ho, P.~Arunachalam, M.~Somaiya, and C.-J. Lin, ``Product title
  classification versus text classification,'' \emph{Csie. Ntu. Edu. Tw}, pp.
  1--25, 2012.

\bibitem{banerjee-etal-2019-hierarchical}
\BIBentryALTinterwordspacing
S.~Banerjee, C.~Akkaya, F.~Perez-Sorrosal, and K.~Tsioutsiouliklis,
  ``Hierarchical transfer learning for multi-label text classification,'' in
  \emph{Proceedings of the 57th Annual Meeting of the Association for
  Computational Linguistics}.\hskip 1em plus 0.5em minus 0.4em\relax Florence,
  Italy: Association for Computational Linguistics, Jul. 2019, pp. 6295--6300.
  [Online]. Available: \url{https://aclanthology.org/P19-1633}
\BIBentrySTDinterwordspacing

\bibitem{chained_neural_networks}
\BIBentryALTinterwordspacing
J.~Wehrmann, R.~C. Barros, S.~N.~d. D\^{o}res, and R.~Cerri, ``Hierarchical
  multi-label classification with chained neural networks,'' in
  \emph{Proceedings of the Symposium on Applied Computing}, ser. SAC '17.\hskip
  1em plus 0.5em minus 0.4em\relax New York, NY, USA: Association for Computing
  Machinery, 2017, p. 790–795. [Online]. Available:
  \url{https://doi.org/10.1145/3019612.3019664}
\BIBentrySTDinterwordspacing

\bibitem{10.1145/3302425.3302483}
T.~Lei, Z.~Shi, D.~Liu, L.~Yang, and F.~Zhu, ``A novel cnn-based method for
  question classification in intelligent question answering,'' in
  \emph{Proceedings of the 2018 International Conference on Algorithms,
  Computing and Artificial Intelligence}, ser. ACAI 2018.\hskip 1em plus 0.5em
  minus 0.4em\relax New York, NY, USA: Association for Computing Machinery,
  2018.

\bibitem{BERT}
J.~Devlin, M.~Chang, K.~Lee, and K.~Toutanova, ``{BERT:} pre-training of deep
  bidirectional transformers for language understanding,'' \emph{CoRR}, vol.
  abs/1810.04805, 2018.

\bibitem{MachinT}
L.~Tan, M.~Y. Li, and S.~Kok, ``E-commerce product categorization via machine
  translation,'' \emph{ACM Trans. Manage. Inf. Syst.}, vol.~11, no.~3, 2020.

\bibitem{sinha2018hierarchical}
K.~Sinha, Y.~Dong, J.~C.~K. Cheung, and D.~Ruths, ``A hierarchical neural
  attention-based text classifier,'' in \emph{Proceedings of the 2018
  Conference on Empirical Methods in Natural Language Processing}, 2018, pp.
  817--823.

\bibitem{zhou-etal-2020-hierarchy}
\BIBentryALTinterwordspacing
J.~Zhou, C.~Ma, D.~Long, G.~Xu, N.~Ding, H.~Zhang, P.~Xie, and G.~Liu,
  ``Hierarchy-aware global model for hierarchical text classification,'' in
  \emph{Proceedings of the 58th Annual Meeting of the Association for
  Computational Linguistics}.\hskip 1em plus 0.5em minus 0.4em\relax Online:
  Association for Computational Linguistics, Jul. 2020, pp. 1106--1117.
  [Online]. Available: \url{https://aclanthology.org/2020.acl-main.104}
\BIBentrySTDinterwordspacing

\bibitem{lu-etal-2020-multi}
\BIBentryALTinterwordspacing
J.~Lu, L.~Du, M.~Liu, and J.~Dipnall, ``Multi-label few/zero-shot learning with
  knowledge aggregated from multiple label graphs,'' in \emph{Proceedings of
  the 2020 Conference on Empirical Methods in Natural Language Processing
  (EMNLP)}.\hskip 1em plus 0.5em minus 0.4em\relax Online: Association for
  Computational Linguistics, Nov. 2020, pp. 2935--2943. [Online]. Available:
  \url{https://aclanthology.org/2020.emnlp-main.235}
\BIBentrySTDinterwordspacing

\bibitem{Frnkranz2008MultilabelCV}
J.~F{\"u}rnkranz, E.~H{\"u}llermeier, E.~L. Menc{\'i}a, and K.~Brinker,
  ``Multilabel classification via calibrated label ranking,'' \emph{Machine
  Learning}, vol.~73, pp. 133--153, 2008.

\bibitem{10.5555/2999792.2999959}
T.~Mikolov, I.~Sutskever, K.~Chen, G.~Corrado, and J.~Dean, ``Distributed
  representations of words and phrases and their compositionality,'' in
  \emph{Proceedings of the 26th International Conference on Neural Information
  Processing Systems - Volume 2}, ser. NIPS’13.\hskip 1em plus 0.5em minus
  0.4em\relax Red Hook, NY, USA: Curran Associates Inc., 2013, p. 3111–3119.

\bibitem{pennington-etal-2014-glove}
J.~Pennington, R.~Socher, and C.~Manning, ``{G}love: Global vectors for word
  representation,'' in \emph{Proceedings of the 2014 Conference on Empirical
  Methods in Natural Language Processing ({EMNLP})}.\hskip 1em plus 0.5em minus
  0.4em\relax Doha, Qatar: Association for Computational Linguistics, Oct.
  2014, pp. 1532--1543.

\bibitem{pagliardini2017unsupervised}
M.~Pagliardini, P.~Gupta, and M.~Jaggi, ``Unsupervised learning of sentence
  embeddings using compositional n-gram features,'' \emph{arXiv preprint
  arXiv:1703.02507}, 2017.

\bibitem{reimers-gurevych-2019-sentence}
N.~Reimers and I.~Gurevych, ``Sentence-{BERT}: Sentence embeddings using
  {S}iamese {BERT}-networks,'' in \emph{Proceedings of the 2019 Conference on
  Empirical Methods in Natural Language Processing and the 9th International
  Joint Conference on Natural Language Processing (EMNLP-IJCNLP)}.\hskip 1em
  plus 0.5em minus 0.4em\relax Hong Kong, China: Association for Computational
  Linguistics, Nov. 2019.

\bibitem{cer-etal-2018-universal}
D.~Cer, Y.~Yang, S.-y. Kong, N.~Hua, N.~Limtiaco, R.~St.~John, N.~Constant,
  M.~Guajardo-Cespedes, S.~Yuan, C.~Tar, B.~Strope, and R.~Kurzweil,
  ``Universal sentence encoder for {E}nglish,'' in \emph{Proceedings of the
  2018 Conference on Empirical Methods in Natural Language Processing: System
  Demonstrations}.\hskip 1em plus 0.5em minus 0.4em\relax Brussels, Belgium:
  Association for Computational Linguistics, Nov. 2018, pp. 169--174.

\bibitem{compositionality1}
\BIBentryALTinterwordspacing
H.~Bhathena, A.~Willis, and N.~Dass, ``Evaluating compositionality of sentence
  representation models,'' in \emph{Proceedings of the 5th Workshop on
  Representation Learning for NLP}.\hskip 1em plus 0.5em minus 0.4em\relax
  Online: Association for Computational Linguistics, Jul. 2020, pp. 185--193.
  [Online]. Available: \url{https://aclanthology.org/2020.repl4nlp-1.22}
\BIBentrySTDinterwordspacing

\bibitem{compositionality2}
\BIBentryALTinterwordspacing
A.~Ettinger, A.~Elgohary, C.~Phillips, and P.~Resnik, ``Assessing composition
  in sentence vector representations,'' in \emph{Proceedings of the 27th
  International Conference on Computational Linguistics}.\hskip 1em plus 0.5em
  minus 0.4em\relax Santa Fe, New Mexico, USA: Association for Computational
  Linguistics, Aug. 2018, pp. 1790--1801. [Online]. Available:
  \url{https://aclanthology.org/C18-1152}
\BIBentrySTDinterwordspacing

\bibitem{mazare-etal-2018-training}
\BIBentryALTinterwordspacing
P.-E. Mazar{\'e}, S.~Humeau, M.~Raison, and A.~Bordes, ``Training millions of
  personalized dialogue agents,'' in \emph{Proceedings of the 2018 Conference
  on Empirical Methods in Natural Language Processing}.\hskip 1em plus 0.5em
  minus 0.4em\relax Brussels, Belgium: Association for Computational
  Linguistics, Oct.-Nov. 2018, pp. 2775--2779. [Online]. Available:
  \url{https://aclanthology.org/D18-1298}
\BIBentrySTDinterwordspacing

\bibitem{biencoder2}
\BIBentryALTinterwordspacing
E.~Dinan, V.~Logacheva, V.~Malykh, A.~Miller, K.~Shuster, J.~Urbanek, D.~Kiela,
  A.~Szlam, I.~Serban, R.~Lowe, S.~Prabhumoye, A.~W. Black, A.~Rudnicky,
  J.~Williams, J.~Pineau, M.~Burtsev, and J.~Weston, ``The second
  conversational intelligence challenge (convai2),'' 2019. [Online]. Available:
  \url{https://arxiv.org/abs/1902.00098}
\BIBentrySTDinterwordspacing

\bibitem{randomneg}
\BIBentryALTinterwordspacing
J.-T. Huang, A.~Sharma, S.~Sun, L.~Xia, D.~Zhang, P.~Pronin, J.~Padmanabhan,
  G.~Ottaviano, and L.~Yang, ``Embedding-based retrieval in facebook search,''
  in \emph{Proceedings of the 26th ACM SIGKDD International Conference on
  Knowledge Discovery and Data Mining}, ser. KDD '20.\hskip 1em plus 0.5em
  minus 0.4em\relax New York, NY, USA: Association for Computing Machinery,
  2020, p. 2553–2561. [Online]. Available:
  \url{https://doi.org/10.1145/3394486.3403305}
\BIBentrySTDinterwordspacing

\bibitem{hardneg1}
\BIBentryALTinterwordspacing
L.~Xiong, C.~Xiong, Y.~Li, K.-F. Tang, J.~Liu, P.~Bennett, J.~Ahmed, and
  A.~Overwijk, ``Approximate nearest neighbor negative contrastive learning for
  dense text retrieval,'' 2020. [Online]. Available:
  \url{https://arxiv.org/abs/2007.00808}
\BIBentrySTDinterwordspacing

\bibitem{hardneg2}
\BIBentryALTinterwordspacing
K.~Guu, K.~Lee, Z.~Tung, P.~Pasupat, and M.-W. Chang, ``Realm:
  Retrieval-augmented language model pre-training,'' 2020. [Online]. Available:
  \url{https://arxiv.org/abs/2002.08909}
\BIBentrySTDinterwordspacing

\bibitem{fast_transformers}
\BIBentryALTinterwordspacing
A.~Vyas, A.~Katharopoulos, and F.~Fleuret, ``Fast transformers with clustered
  attention,'' 2020. [Online]. Available:
  \url{https://arxiv.org/abs/2007.04825}
\BIBentrySTDinterwordspacing

\bibitem{softmax}
\BIBentryALTinterwordspacing
B.~Gao and L.~Pavel, ``On the properties of the softmax function with
  application in game theory and reinforcement learning,'' 2017. [Online].
  Available: \url{https://arxiv.org/abs/1704.00805}
\BIBentrySTDinterwordspacing

\bibitem{frome2013devise}
A.~Frome, G.~S. Corrado, J.~Shlens, S.~Bengio, J.~Dean, M.~Ranzato, and
  T.~Mikolov, ``Devise: A deep visual-semantic embedding model,'' in
  \emph{Advances in neural information processing systems}, 2013, pp.
  2121--2129.

\bibitem{snell2017prototypical}
J.~Snell, K.~Swersky, and R.~Zemel, ``Prototypical networks for few-shot
  learning,'' in \emph{Advances in neural information processing systems},
  2017, pp. 4077--4087.

\end{thebibliography}

\begin{IEEEbiography}[{\includegraphics[width=1in,height=1.25in,clip,keepaspectratio]{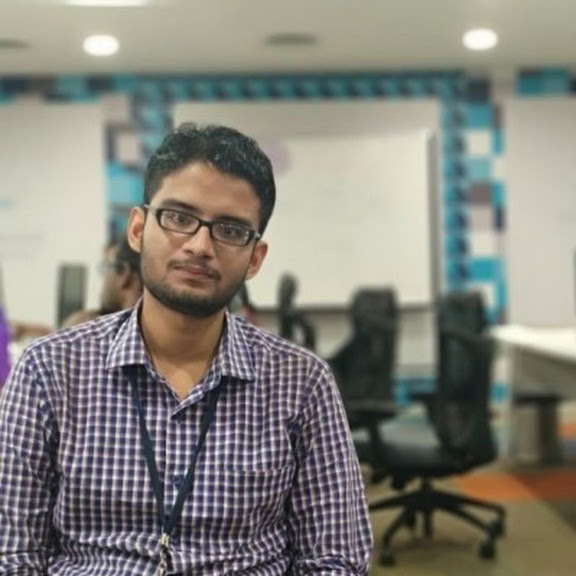}}]{Venktesh V} is a PhD student in the Department of Computer Science and Engineering in IIIT-Delhi and a PM fellow (SERB-FICCI). His interests are in neural Information Retrieval and Natural Language Processing. 
\end{IEEEbiography}

\begin{IEEEbiography}[{\includegraphics[width=1in,height=1.25in,clip,keepaspectratio]{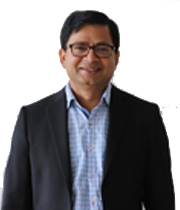}}]{Mukesh Mohania} is a Professor in the Department of Computer Science and Engineering and Dean of Innovation, Research and Development at IIIT-Delhi. He has published more than 120 Research papers in reputed International Conferences and Journals and has more than 50 granted patents. His research interests are on Information (structured and unstructured data) integration, Natural Language Processing, AI based entity analytics, and big data analytics and applications. 
\end{IEEEbiography}

\begin{IEEEbiography}[{\includegraphics[width=1in,height=1.25in,clip,keepaspectratio]{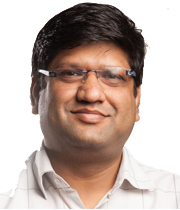}}]{Vikram Goyal} is a professor in the Department of Computer Science and Engineering in IIIT-Delhi. He has many publications in reputed conferences and referred journals. He is the director of TiH Anubhuti (IIITD) and Program Director of PG Diploma in Data Science and AI. He is also a member of Infosys Centre for AI at IIITD. His research interests are in information Retrieval, Data Mining, Databases, and Spatial Data Analytics
\end{IEEEbiography}

%\vfill


\end{document}